\documentclass{article}
\usepackage{spconf,amsmath,graphicx}
\usepackage{multirow,amssymb}
\usepackage{cite}
\usepackage{xcolor}
\usepackage{tikz}
\usepackage{lipsum}

\newcommand\copyrighttext{%
  \footnotesize \copyright  2022 IEEE.  Personal use of this material is permitted.  Permission from IEEE must be obtained for all other uses, in any current or future media, including reprinting/republishing this material for advertising or promotional purposes, creating new collective works, for resale or redistribution to servers or lists, or reuse of any copyrighted component of this work in other works.}
\newcommand\mycopyrightnotice{%
\begin{tikzpicture}[remember picture,overlay]
\node[anchor=south,yshift=10pt] at (current page.south) {\fbox{\parbox{\dimexpr\textwidth-\fboxsep-\fboxrule\relax}{\copyrighttext}}};
\end{tikzpicture}%
}

\title{Nested Multiple Instance Learning with Attention Mechanisms}
\name{Saul Fuster, Trygve Eftestøl, Kjersti Engan\thanks{This research has received funding from the European Union's Horizon 2020 research and innovation program under grant agreements 860627 (CLARIFY).}}
\address{Dept. of Electrical Engineering and Computer Science, University of Stavanger, Norway \\ \{saul.fusternavarro, trygve.eftestol, kjersti.engan\}@uis.no}
\begin{document}
\ninept
\maketitle
\mycopyrightnotice
\begin{abstract}
Strongly supervised learning requires detailed knowledge of truth labels at instance levels, and in many machine learning applications this is a major drawback. Multiple instance learning (MIL) is a popular weakly supervised learning method where truth labels are not available at instance level, but only at bag-of-instances level. However, sometimes the nature of the problem requires a more complex description, where a nested architecture of bag-of-bags at different levels can capture underlying relationships, like similar instances grouped together. Predicting the latent labels of instances or inner-bags might be as important as predicting the final bag-of-bags label but is lost in a straightforward nested setting.  We propose a Nested Multiple Instance with Attention (NMIA) model architecture combining the concept of nesting with attention mechanisms. We show that NMIA performs as conventional MIL in simple scenarios and can grasp a complex scenario providing insights to the latent labels at different levels.
\end{abstract}
\section{Introduction}
\label{sec:intro}


Multiple instance learning (MIL) is a learning method where several elements, called instances, are individually unlabelled \cite{Maron1997AFF}. However, a label exists for a group of such instances, called a bag. MIL is an example of weakly supervised learning methods. Usually, in MIL, a bag label is defined as positive if at least one of the instances is positive and negative if all instances are negative. We will refer to this as the MIL assumption. In MIL, the multiple instances feature representation are aggregated into a single representation used in a supervised learning setup with the weak label.

Weakly supervised learning is suited for medical applications where patient-based, clinical labels are known, whereas detailed localized annotations in recorded biosignals or images often are unavailable. An example is in digital pathology, where histopathological whole slide images (WSI) are high-resolution digital files of scanned microscopic tissue sections from biopsies. WSI are of gigantic dimensions; hence they are typically referred to as gigapixel images, and processing them at once is infeasible, thus the images are divided in patches. Furthermore, annotating such a large image in detail is very cumbersome and time-consuming due to the size of the image and tumours being a heterogeneous disease challenging to diagnose. Therefore, the number of annotations is limited and often pathology datasets rely exclusively on clinicopathological information, where each image patch can be considered an instance. If a tissue section is cancerous, the positive (i.e. cancerous) instances would typically be localized in one or several regions that share similar cellular features, and it would typically not be single positive instances surrounded by negative ones. A conventional weakly supervised model as MIL would not perceive this sense of location since all instances are grouped and features aggregated under the same bag. One way to overcome this is to introduce Nested MIL (NMIL), allowing to group the extracted patches from regions separately while training the model on a weak label.

Even if such a nested system can perform well, the knowledge of which individual instances or inner-bags are the most impactful will be lost with a straightforward setup of MIL. This results in a low degree of explainability and applicability for interpreting a particular prediction, which has been one of the major focus in deep learning recently \cite{Gilpin2018ExplainingEA,Xie2020ExplainableDL}. Attention mechanisms can be integrated in a weakly supervised model architecture to provide a degree of explainability at instance level. Such models are also showing comparable or improved performance on their bag predictions \cite{Ilse2018AttentionbasedDM}. Attention scores reveal how input features are weighted and can be visualized as a magnitude value of the instance significance \cite{Li2020ArnetAR,Guo2021ChannelAR,Xu2019AnIA,Hong2020WeaklyLA}.

Our proposed model architecture, Nested Multiple Instance with Attention (NMIA), that overcomes the intricate relationship between instance and bag labels with nesting and offers a high degree of interpretability using attention mechanisms. To validate the idea of finding bags with multiple positive instances, simulating a  region of intestest (ROI) in a WSI, we make use of PCAM \cite{Bejnordi2017DiagnosticAO,Veeling2018RotationEC}, an image dataset consisting of patches extracted from WSI of lymph node sections. Moreover, to prove that nesting can perform in other domains, the MNIST dataset is used, a classical image dataset consisting of handwritten digits \cite{LeCun1998GradientbasedLA}.

\section{Related work}
\label{sec:relatedwork}

Several applications have been developed that use MIL for overcoming the lack of annotated regions of interest in an image \cite{Amores2013MultipleIC,Carbonneau2018MultipleIL}. Chen et al. \cite{Chen2021AnAW} train end-to-end feeding an entire WSI in a strong supervision manner, adopting a variant of MIL. To provide further intuition into the composition of the bags and the relevance that individual instances carry in the classification, Chikontwe et al. \cite{Chikontwe2020MultipleIL} propose a center loss that characterizes intra-class variations by minimizing the distance among instances from the same class. Li et al. \cite{Li2021DualstreamMI} propose a dual-stream architecture to learn instance and bag classifiers at once, where the first instance would be an instance classifier and the second stream aggregates the instances into a bag embedding to feed to a bag classifier. Also, He et al. \cite{He2020ClusteringbasedMI} use a clustering-based strategy to obtain hidden structure information in the feature space to discover positive instances. These methods work well under the MIL assumption but would not necessarily understand more \emph{complex scenarios}, such as region-based analysis of WSI, detection of sequential events in a time-series signal, natural language processing of blocks of text, among others.

The concept of bag-of-bags is first seen in an application for prostate cancer detection using magnetic resonance images \cite{Khalvati2016BagOB}. The concept of a nested architecture for MIL is later introduced by Tibo et al. \cite{Tibo2017ANA,Tibo2020LearningAI}, where they present the use of bag layers to aggregate instance-level representation into a bag-level representation. This implementation, however, remains opaque since its architecture does not offer interpretability for which instances or inner-bags contributed the most to the final prediction. Adding nesting exacerbates this issue since the gap between the instances and the final weak label is even more prominent. 

Attention mechanisms were introduced into weakly supervised methods to give insight into the model's decision making and ability to pick out instances of interest \cite{Lu2021DataEA,Li2021AMM,Pirovano2021AutomaticFS,Ilse2018AttentionbasedDM,Sharma2021ClustertoConquerAF,Xie2019VisualEA}. A trainable attention mechanism identifies the instances that have a more significant influence in making a positive prediction. This is self-enforcing by using the attention scores to strengthen the instance representation before aggregation. The final prediction is made using the aggregated representation.

In this work, we propose Nested Multiple Instance with Attention (NMIA), a novel model architecture for weakly supervised learning methods on structured data. We are further developing the concept of nesting for weakly supervised learning methods and combining it with attention mechanisms for interpretability.

\section{Methodology}
\label{sec:methodology}

\subsection{Nested Multiple Instance Learning}
\label{ssec:nmil}
Multiple instance learning (MIL) is a weakly supervised method trained in a supervised manner considering outermost bags and their corresponding labels. For the conventional MIL binary setting, a dataset $\mathcal{X}, \mathcal{Y} = \left \{ (\mathbf{X}^i,y^i), \forall i=1,...,N \right \}$ is formed of pairs of sample sets $\mathbf{X}$ and their corresponding labels $y$, where $i$ denotes the current sample for a total of $N$ samples. A sample $\mathbf{X}$ consists of a bag of instances $\mathbf{x}_l$:
\begin{equation}
    \mathbf{X}= \left \{ \mathbf{x}_l, \forall l=1,...,L \right \}
    \label{eq:boi}
\end{equation}

where $L$ is the number of instances in the bag. In order to obtain an instance representation/embedding from input data, for example an image patch, we make use of a feature extractor. An image feature extractor $G_f : \bar{\mathbf{x}} \to \mathbf{x}$ is tipically a convolutional neural network which maps an image patch $\bar{\mathbf{x}}$ into a feature vector $\mathbf{x}$. Instance representations $\mathbf{x}$ from a bag $\textbf{X}$ are aggregated to form a bag representation using an aggregation function $\Xi$, which can, for example, be replaced by either the $mean$ or $max$ operators.

A label $y \in \left \{ 0,1 \right \}$ is associated with the bag $\mathbf{X}$. Although each instance might be associated with a label $y_l$, they are generally unknown; hence only bag labels are used during training. At the inference stage, a test set might be associated with labels both at the bag and instance level to provide performance metrics. Under the conventional binary classification MIL assumption, a bag label is positive with the single presence of a positive instance. Then, the model learns comparing $y$ with the prediction $\hat{y}$, computed by the bag classifier $\Theta_c$.
\begin{equation}
    y = \underset{l}{max}\{y_l\}
    \label{eq:binary}
\end{equation}

\begin{equation}
    \hat{y} = \Theta_c(\Xi(\textbf{X}))
    \label{eq:pred}
\end{equation}

\begin{figure}[htbp]
\centerline{\includegraphics[width=\columnwidth, height=1.8125in]{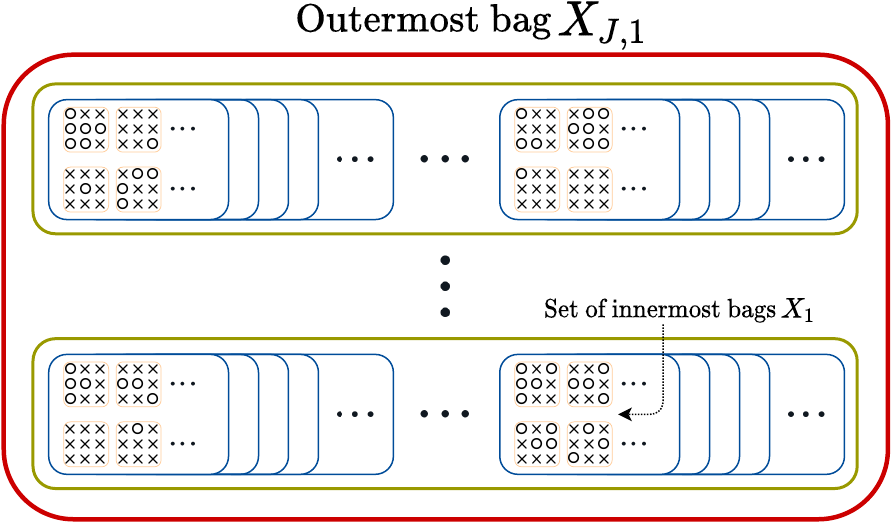}}
\caption{Nested bag-of-bags. Instances (crosses and circles) are drawn into the innermost bags to form sets which are recursively grouped finally forming the outermost bag.}
\label{fig:nestedconcept}
\end{figure}

In contrast with conventional MIL, NMIL setting consists of levels of bags within bags where only the innermost bags contain instances. In Figure \ref{fig:nestedconcept} the idea of grouping instances in inner-bags, inner-bags in larger bags, and finally in one outermost bag is illustrated.

Let $J$ denote the number of nested levels, $K_{j}$ number of bags at level $j$ and $L_{j,k}$ denote the number of instances or bags in a bag $k$. A set $\mathbf{X}_j$ contains a set of inner-bags $\mathbf{X}_{j,k}$:
\begin{equation}
    \mathbf{X}_{j} = \left \{ \mathbf{X}_{j,k}, \forall k=1,...,K_{j} \right \}
    \label{eq:bob}
\end{equation}

for $j$ defining the current nesting level up to $J$ levels of nesting. For $J=1$, NMIL with one level of nesting, the NMIL notation corresponds to the ordinary MIL notation described in Eq. (\ref{eq:boi}). The number of inner-bags $K_{j}$ can vary from level to level. For a given inner-bag $k$ at level $j$, where $l$ defines the instance number up to $L_{j,k}$, a bag of instances $\mathbf{X}_{j,k}$ is expressed as:
\begin{equation}
    \mathbf{X}_{j,k}= \left \{ \mathbf{x}_{j,k,l}, \forall l=1,...,L_{j,k} \right \}
    \label{eq:innermost}
\end{equation}


By latent labels, we refer to the actual, but in general unknown, labels of an instance or inner-bag. At the training stage, however, these are necessary to determine the final bag-of-bags weak label $y$. For simplicity in the experiments section, we will refer to an instance latent label $y_{j,k,l}$ as $y^j_l$ to describe the latent label of an instance $l$ in a level $j$, omitting bag index $k$.

\subsection{Attention Mechanism}
\label{ssec:attention}
\begin{figure}[htbp]
\centerline{\includegraphics[trim={0 0 0.65cm 0},clip,scale=0.285]{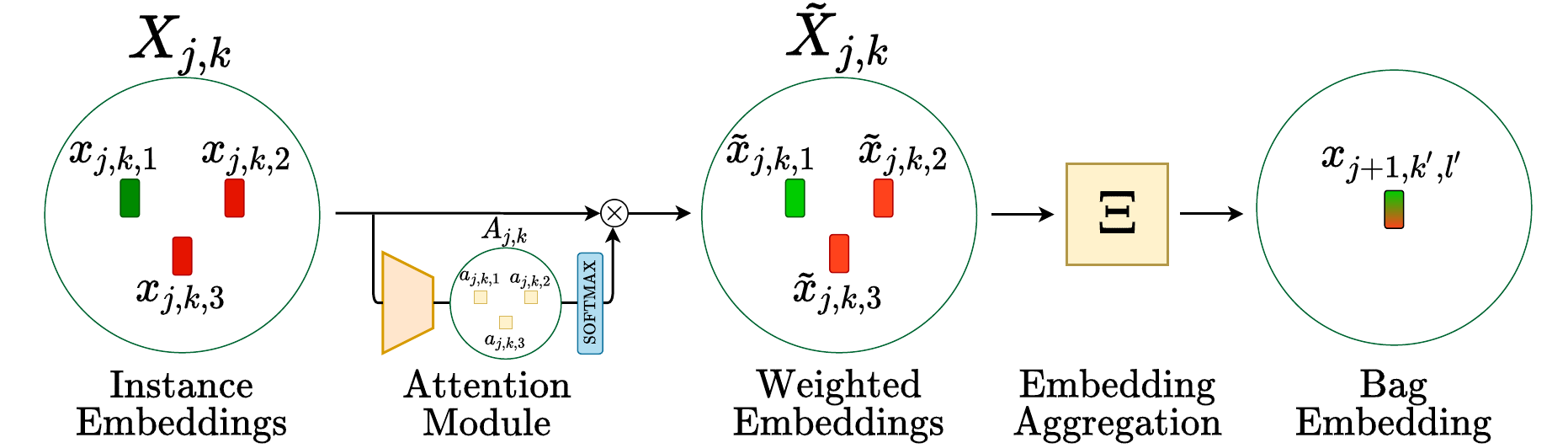}}
\caption{MIA block. Input embeddings are fed into the attention module to compute attention scores. Those scores are used to compute weighted representations of instance embeddings. Then, these weighted embeddings are aggregated to create a final bag embedding.}
\label{fig:MILblock}
\end{figure}

According to Ilse et al \cite{Ilse2018AttentionbasedDM}, a multiple instance block is constructed using an embedding-level approach to obtain a bag-level representation from the instances within, as depicted in Figure \ref{fig:MILblock}. The proposed attention module's input corresponds to low-dimensional embeddings. These are generated by a feature extractor or by bag embedding representations of previous levels. This module observes those embeddings and computes attention scores that leverage the meaningfulness of the features extracted for the given task. Finally, the attention scores are aggregated to obtain the bag representation.


For ease of exposition, we define indexes $\delta = (j,k,l)$, ${{\delta}'} = (j+1,{k}',{l}')$, $\gamma = (j,k)$. We omit the use of training sample index $i$. An attention score $a_\delta$ for a given input embedding $x_\delta$ is calculated as:
\begin{equation}
    a_{\delta} =  \mathrm{exp}\{ \mathbf{w}^{\top} (\mathrm{tanh}(\mathbf{Vx_{\delta}^{\top}}) \odot \mathrm{sigm}(\mathbf{Ux_{\delta}^{\top})})\}
\label{eq:attscore}
\end{equation}

where $\mathbf{w}\in\mathbb{R}^{L\times 1}$, $\mathbf{V}\in\mathbb{R}^{L\times M}$ and $\mathbf{U}\in\mathbb{R}^{L\times M}$ are trainable parameters and $\odot$ is an element-wise multiplication. Furthermore, the hyperbolic tangent $\mathrm{tanh}(\cdot)$ and sigmoid $\mathrm{sigm}(\cdot)$ are included to introduce non-linearity for learning complex applications. Then, attention scores $a_\delta$ are normalized into $\tilde{a}_\delta$ to ensure that the sum of the components of the attention scores vector is 1, as this makes it possible to have variable bag sizes. Note that the unnormalized $a_{\delta}$ would better reflect the attention score directly, and we use that for visualization in the experiments. 

\begin{equation}
    \tilde{a}_\delta =  \frac{a_\delta} {\sum_{l=1}^{L_{\gamma}}a_\delta}
\label{eq:attscoreden}
\end{equation}

Finally, the aggregation function $\Xi$ transforms a leveraged bag of embeddings $\mathbf{\tilde{X}_{\gamma}}$ to obtain a bag representation $\mathbf{x_{{\delta}'}}$ as: 

\begin{equation}
    \mathbf{x_{{\delta}'}} = \Xi(\mathbf{\tilde{X}_{\gamma}}) = \Xi(\{\tilde{a}_\delta \cdot \mathbf{x_{\delta}}, \forall l=1,...,L_{\gamma}\})
\label{eq:insagg}
\end{equation}

\begin{figure}[h]
\centerline{\includegraphics[trim={-0.75cm 0.55cm -0.50cm 0.1cm},clip,scale=0.755]{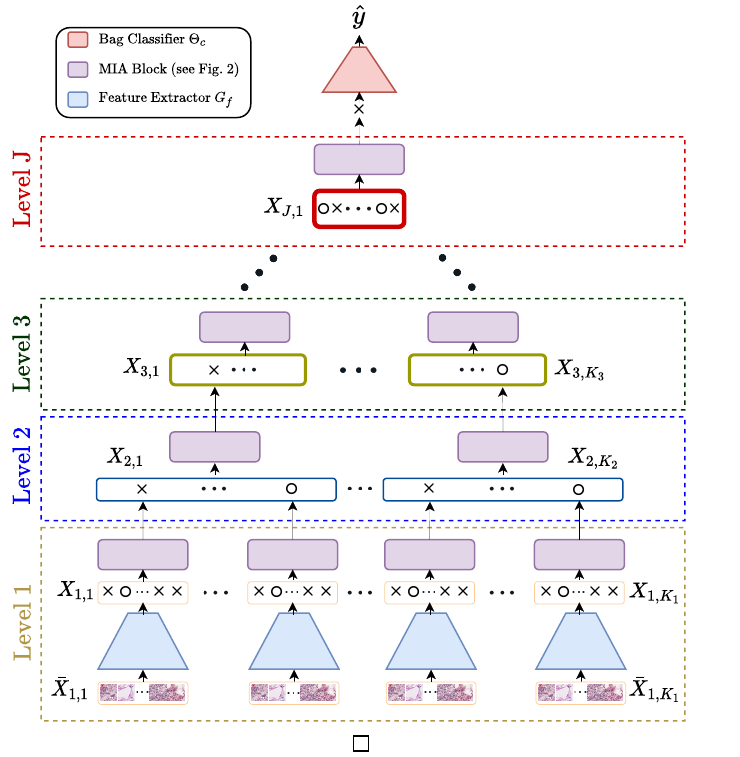}}
\caption{NMIA model architecture. The feature extractor $G_f$ projects all instances into low-dimensional embeddings. Consecutive MIA blocks aggregate deeper levels into more superficial representations. Finally, a bag-of-bags embedding is fed to the classifier $\Theta_c$ for obtaining a bag prediction $\hat{y}$.}
\label{fig:NMILmodel}
\end{figure}

\subsection{Model Architecture}
\label{sec:architecture}

The proposed neural network architecture NMIA combines a recursive bag processing of low-dimensional embeddings and attention mechanisms. The multiple instance with attention (MIA) block computes attention scores for each input embedding and aggregates them to create an embedding that represents the bag's contents, see Figure \ref{fig:MILblock}. Instances are transformed into embeddings, weighted and aggregated into a bag-of-bags embedding after passing through the MIA blocks, see Figure \ref{fig:NMILmodel}. Finally, a classifier predicts the label $\hat{y}$ for the sample $\textbf{X}$.

\section{EXPERIMENTAL SETUP}
\label{sec:experiments}

Several experiments were carried out to show the usefulness of NMIA compared to multiple instance (MI) architectures under different types of data and tasks. Also, we show the importance of attention-based models to obtain further interpretability from meaningful instances and compare the performance to traditional aggregation techniques. Two image datasets were used: MNIST \cite{LeCun1998GradientbasedLA} and PCAM \cite{Bejnordi2017DiagnosticAO,Veeling2018RotationEC}. MNIST consists of a training and test set of 60,000 and 10,000 examples, respectively. PCAM consists of 327,680 patches, where each patch is annotated in a binary manner to indicate the presence of metastatic tissue. All models were trained using stochastic gradient descent (SGD) optimizer, binary cross-entropy loss function and early stopping. A custom convolutional neural network and VGG16 were used as feature extractors $G_f$ for MNIST and PCAM, respectively. The models were trained and tested with 20,000 and 5,000 bag-of-bags, respectively, constructed by randomly extracting instances within the class of desired instance labels. All models are implemented in Python 3.6 using Tensorflow machine learning library \cite{Abadi2016TensorFlowAS}. The code is publicly available on our GitHub repository (https://bit.ly/3JGDibl).


We have conducted three experiments as follows. In the first two, we have considered the positive instance class $y^1_{l^+}$ as the digit 9 for MNIST and tiles with metastatic tissue for PCAM. Dataset samples were arranged to form a 2-level setting. The third experiment was carried out exclusively for MNIST in a 3-level setting. Latent labels $y^j_l$ introduced in intermediate levels are formulated only to obtain the resulting bag-of-bags labels, and to be compared with attention scores to evaluate if we find the correct latent labels. All models are trained entirely on weak bag-of-bags labels $y$.

\textit{\textbf{Exp1:}} A dataset is constructed following the MIL assumption described in Eq. (\ref{eq:binary}). In this assumption, there is nothing to gain in using NMIA, but we want to show that the NMIA architecture is flexible and the model will perform comparably to MI models. A conventional MI model both with and without attention is compared to 2-level NMI models with random grouping of the bags at the first level.
    
\textit{\textbf{Exp2:}} A dataset is constructed such that at least two instances from the same inner-bag have to be positive for the weak bag-of-bags label to be positive, as described in Eq. (\ref{eq:2level_mnist}). This is motivated from region-based analysis of medical images, where typically an object belonging to a positive class is located in a specific region and not scattered across the entire image. Therefore, this particular positive region of the image will contain several positive instances. Regions containing few positives are regarded as noise or misclassified instances, and they should not be reflected in the overall prediction. 
\begin{equation}
\begin{matrix}

\mathbf{1^{st}}
&
\mathbf{2^{nd}}
\\
y^2_l =
\left\{\begin{matrix}
0, \;\; \#y^1_{l^+} \leq \, 1
\\  
1, \;\; \#y^1_{l^+} \, > 1 
\end{matrix}\right.
 &
y = 
\left\{\begin{matrix}
0, \;\; y^2_l \notin 1 
\\  
1, \;\; y^2_l \in 1   
\end{matrix}\right.

\end{matrix}
\label{eq:2level_mnist}
\end{equation}

\begin{figure}[!htbp]
\centerline{\includegraphics[trim={-0.35cm 0.2cm 0.6cm 0},clip,scale=0.51]{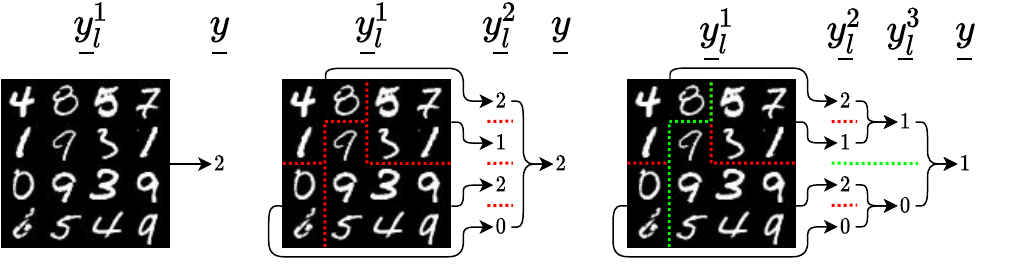}}
\setlength{\belowcaptionskip}{-8pt}
\caption{From left to right, 1, 2 and 3-level partitioning. Red dotted lines separate bags-of-instances at a second level, while greens at a third level. The task of \textit{Exp3} is to find out if a second-level bags contains at least one first-level bag with odd numbers but no first-level bags containing only even numbers.}
\label{fig:3images}
\end{figure}

\textit{\textbf{Exp3:}} Here, we want to find a ROI that contains a bag of odd numbers and not one of even numbers and a 2-level solution is not enough to overcome this task; hence three levels of nesting are required. A region in the image is considered 0 if all instances are even numbers, 1 if they are odd numbers and 2 if there is a mix. In Figure \ref{fig:3images}, an example is shown where the entire image is one region, the image partitioned in regions and regions with sub-regions, respectively. A 3-level partitioning can be used to construct a dataset reflecting a complex scenario as described in Eq. (\ref{eq:3levels}). To get a final bag label to 1, there has to be second level region that contains a bag of only odd numbers but not any bags of only even numbers. Such outline is impossible to learn using MI models, as proven in \textit{Exp2}; thus, corresponding MI and MIA tests were never carried out.
\begin{equation}
\begin{matrix}
\begin{matrix}
\mathbf{1^{st}}
 &
\mathbf{2^{nd}}
\\
y^2_l = 
\left\{\begin{matrix}
0, \;\; y^1_l \in \left \{ 0,2,4,6,8 \right\}
\\  
1, \;\; y^1_l \in \left \{ 1,3,5,7,9 \right\}
\\
2,\;\; otherwise \;\;\;\;\;\;\;\;\;\;\;
\end{matrix}\right. 
 & 
y^3_l =
\left\{\begin{matrix}
0, \;\; y^2_l \in \{ 0 \cap \bar{1}\} 
\\  
1, \;\; y^2_l \in \{ \bar{0} \cap 1\}  
\\  
2, \;\; otherwise \;\;\;
\end{matrix}\right. 
\end{matrix}
\\
\begin{matrix}
\;\;\;\;\;\;\;\mathbf{3^{rd}} \\
\;\;\;\;\;\;\; y = 
\left\{\begin{matrix}
0, \;\; y^3_l \notin 1 
\\  
1, \;\; y^3_l \in 1 
\end{matrix}\right.
\end{matrix}
\end{matrix}
\label{eq:3levels}
\end{equation}

\section{RESULTS AND DISCUSSION}
\label{sec:results}
F1 scores for the experiments are listed in Table \ref{tbl:bigtable}. MI architecture is compared to MI with attention (MIA), nested MI architecture (NMI) and nested multiple instance architecture with attention, the proposed NMIA architecture. Note the absolute value of the F1 score is dependent on the chosen feature extractor which is not the focus of this paper, rather the relative values between the MI, MIA, NMI an NMIA in simple scenarios (\textit{Exp1}) and more complex scenarios (\textit{Exp2}, \textit{Exp3}) is what we seek to demonstrate.

\textit{Exp1} presents a setup where individual instance latent labels are directly responsible for the resulting bag weak label. Here, we can see that a conventional MI model can perform highly and the choice of architecture does not affect the predictive power of the classifier. Nesting becomes irrelevant when only the individual labels of the instances are meaningful but not their arrangement across inner-bags. However, for \textit{Exp2}, we show that a conventional MI architecture breaks down because it cannot perform or even understand the nature of the task. From the MI model perspective, all instances are at the same level and belong to the same set. \textit{Exp2} and \textit{Exp3} were designed to demonstrate the strength of nesting when the relationship among instances and inner-bags is fundamental for obtaining the final weak label. NMIA can process these subsets independently, hence understanding the relationship among instances and giving insight into which instances and inner-bags are meaningful for the final prediction.

\begin{table}[t]
\caption{F1 scores for experiments on MNIST and PCAM datasets.}\smallskip
\begin{center}
\begin{tabular}{cccc|cc}
                               & \multicolumn{3}{c|}{\textbf{MNIST}}                                                         & \multicolumn{2}{c}{\textbf{PCAM}}                                                          \\
 \cline{2-6} 
                               & \textit{\textbf{Exp1}} & \textit{\textbf{Exp2}} & \textit{\textbf{Exp3}} & \textit{\textbf{Exp1}} & \textit{\textbf{Exp2}} \\ \hline
 \multicolumn{1}{c|}{\textbf{MI}} & 0.929               & 0.345               & N/A               & 0.957               & 0.290               \\
 \multicolumn{1}{c|}{\textbf{MIA}}      & 0.957               & 0.472               & N/A               & 0.973               & 0.286               \\ \hline
 \multicolumn{1}{c|}{\textbf{NMI}} & 0.923               & 0.855               & 0.556               & 0.964               & 0.700               \\
 \multicolumn{1}{c|}{\textbf{NMIA}}      & \textbf{0.959}      & \textbf{0.921}      & \textbf{0.836}               & \textbf{0.978}      & \textbf{0.734}     
\end{tabular}
\end{center}
\label{tbl:bigtable}
\end{table}

\begin{figure}[!htb]
 \begin{minipage}{0.49\columnwidth}
  \includegraphics[width=\linewidth, height=0.95in]{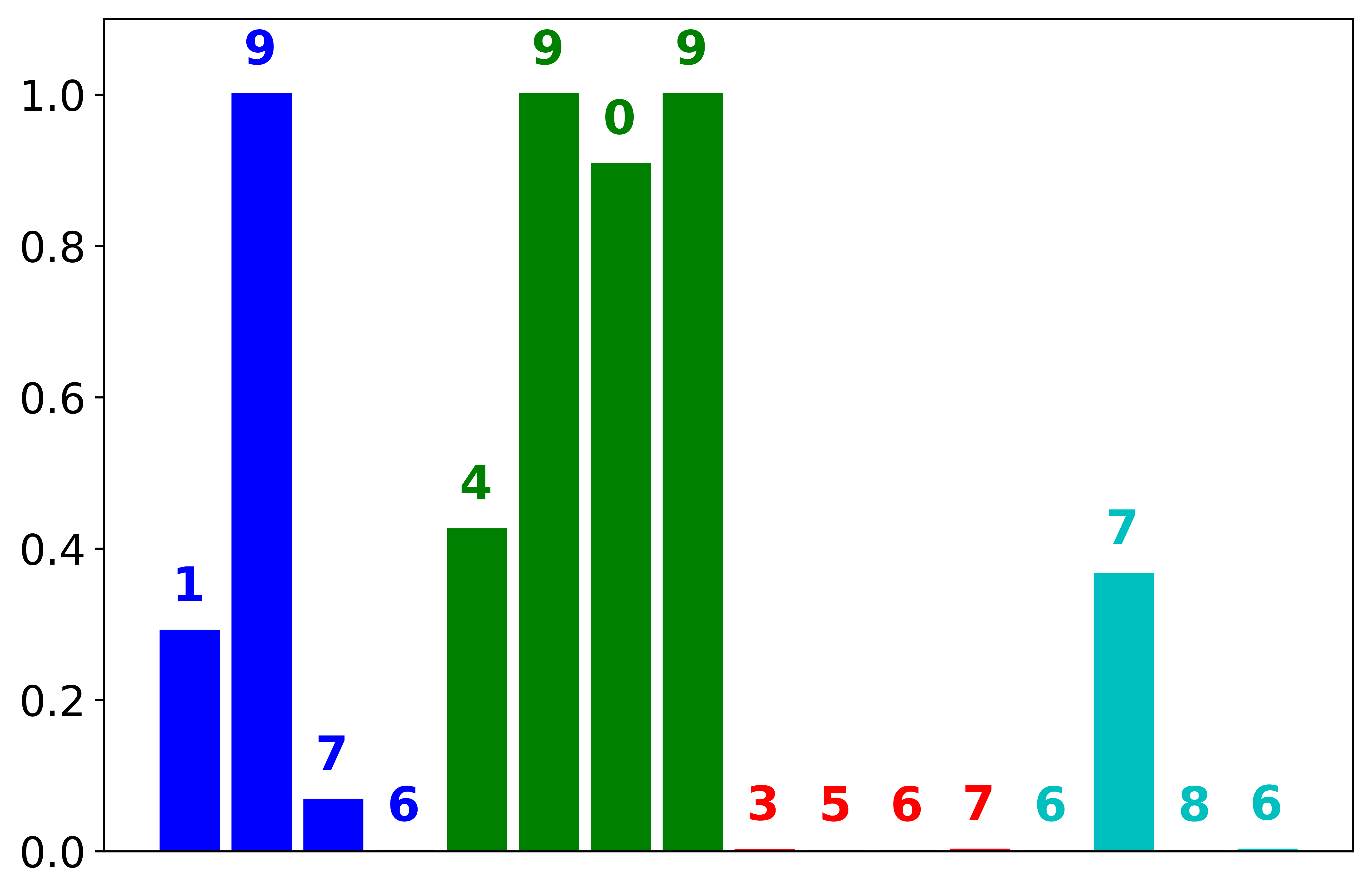}
  \end{minipage}
  \hfill 
    \begin{minipage}{0.49\columnwidth}
  \includegraphics[width=\linewidth, height=0.95in]{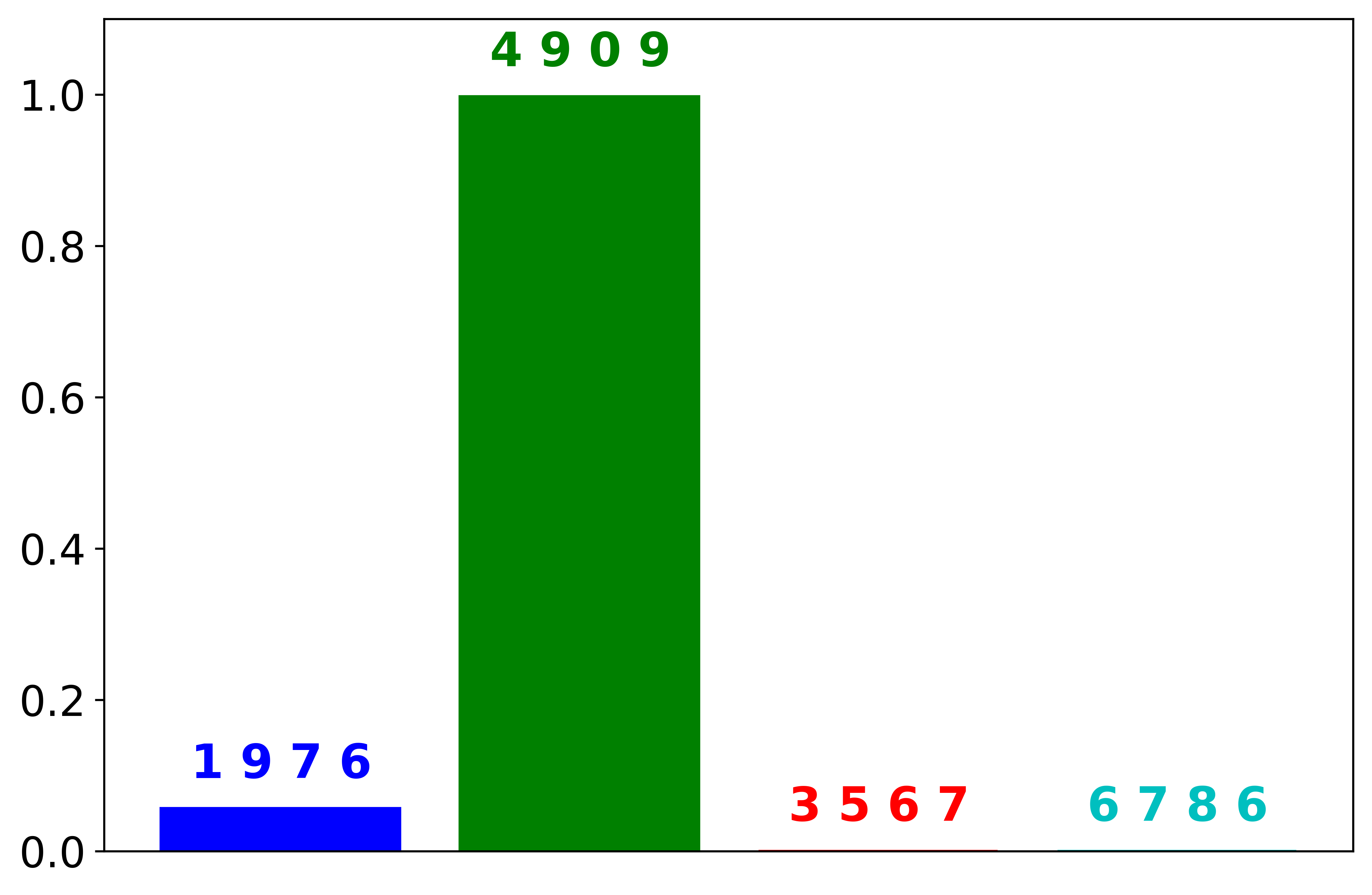}
 \end{minipage}
 \centerline{(a) MNIST} 
 \begin{minipage}{0.49\columnwidth}
  \includegraphics[width=\linewidth, height=0.95in]{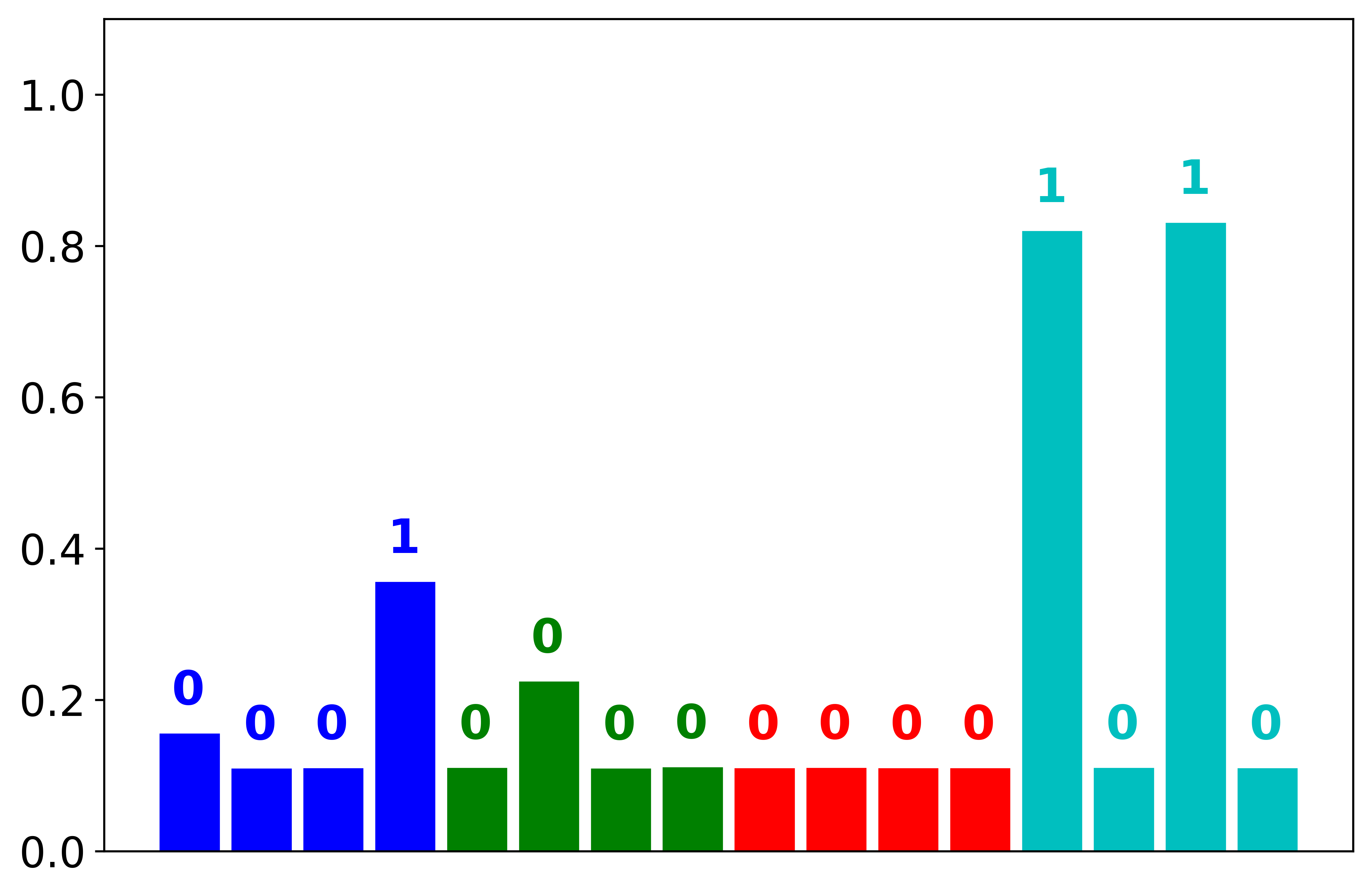}
  \end{minipage}
  \hfill 
    \begin{minipage}{0.49\columnwidth}
  \includegraphics[width=\linewidth, height=0.95in]{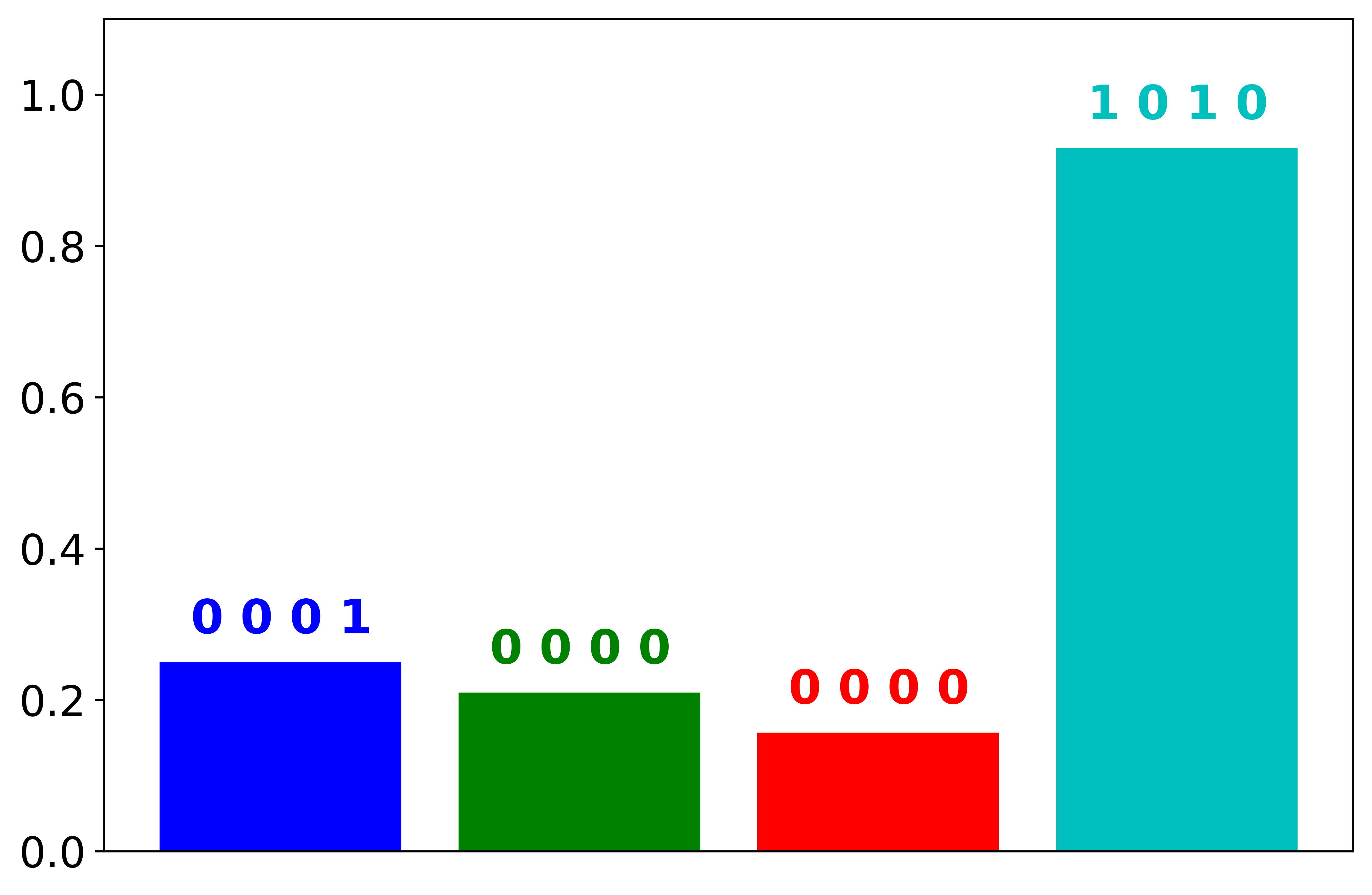}
 \end{minipage}
 \centerline{(b) PCAM}
 \caption{Examples of attention scores $a_\delta$ in test samples from \textit{Exp2} with 2-level NMIA for (a) MNIST and (b) PCAM. A single positive instance in the inner-bag is considered noise whereas two or more should give a positive inner-bag, resulting in a positive bag-of-bags bag. Instances are categorized by colors indicating inner-bag belonging, while digits represent the true label of the instance. Bar plots on the left column show the attention at level 1, while on the right, at level 2. Positive instances obtain the highest attention scores.}
 \label{fig:att_scores}
\end{figure}

Furthermore, implementing a model with an attention mechanism provides an edge over the model that does not, both on performance and interpretability, as shown in Figure \ref{fig:att_scores}. We can observe that the attention mechanism can correctly identify the positive instances at the instance level and recognise the positive inner-bags, distinguishing them from those containing noisy instances. A conventional weakly supervised model would not perceive this sense of location since all instances are encapsulated under the same bag.

\textit{Exp3} further demonstrates NMIAs strengths in a 3-level setup. NMIA model reaches a F1 score of 0.836, thus proving that a nested implementation efficiently handles complex scenarios.

\section{CONCLUSIONS}
\label{sec:conclusions}

In this paper, we have proposed the NMIA architecture for solving applications that simplistic MI architectures cannot, for when dependencies among sets of bags are to be considered. We have presented experiments processing interdependent subsets from images demonstrating the flexibility and improved performance relative to MI. Moreover, NMIA can be used in a wide range of applications due to its flexibility. Finally, implementing an attention mechanism helps identify key instances and inner-bags contained in a set of bags, giving insight into the practical relationship between latent labels and the attention given among different levels. Future research will consider the potential effects of NMIA with attention on a full-scale medical imaging dataset more carefully.

\vfill\pagebreak

\bibliographystyle{IEEEbib}
\bibliography{amores,arnet,bejnordi,carbonneau,chen,chikontwe,dualstream,gilpin,guo,he,hong,ilse,khalvati,lecun,li,lu,maron,pirovano,russakovsky,sharma,tensorflow,tibo,tiboana,veeling,xie,xievisual,xu}

\end{document}